%% file: main.tex
\documentclass[11pt,letterpaper]{article}
\usepackage[T1]{fontenc}
\usepackage{lmodern}
\usepackage[margin=1in]{geometry}
\usepackage{amsmath,amssymb,booktabs,array,graphicx}
\usepackage{algorithm,algpseudocode}
\usepackage[section]{placeins}
\usepackage{changepage}
\usepackage{microtype}
\usepackage[numbers,sort&compress]{natbib}
\usepackage{xurl}
\usepackage[colorlinks=true,linkcolor=blue,citecolor=blue,urlcolor=blue]{hyperref}
\usepackage[font=small,labelfont=bf]{caption}
\newlength{\extralength}
\author{Kautik Mandve\textsuperscript{1}\qquad Dileepa Fernando\textsuperscript{2}\\[0.7em]
\small\textsuperscript{1}Vernon Hills High School, Vernon Hills, Illinois, USA\\
\small\textsuperscript{2}Information Systems Technology and Design\\
\small Singapore University of Technology and Design, Singapore}
\date{2 October 2026}
\title{Explore, Then Commit: Measurement-Efficient Scientific Law Discovery with Language Models}
\hypersetup{pdftitle={Explore, Then Commit: Measurement-Efficient Scientific Law Discovery with Language Models},pdfauthor={Kautik Mandve and Dileepa Fernando}}
\begin{document}
\maketitle
\begin{abstract}
Scientific law discovery requires selecting measurements and converting evidence into a governing equation. We evaluate an explore-then-commit protocol in which a large language model proposes hypotheses, a programmatic planner gathers measurements, and a fresh prompt synthesizes the final law from fixed observations. The protocol combines structured probes, automatic numerical diagnostics, restricted measurement batches, and optional interpreter access. Across 576 NewtonBench trials, we compare eight configurations on 12 physics modules using GPT-4.1-mini and a medium-difficulty GPT-4.1 replication. On medium tasks, interpreter-enabled planners use 8.6 versus 22.5 measurements per trial for GPT-4.1-mini and 8.9 versus 43.0 for GPT-4.1. Their mean magnitude-based root-mean-squared logarithmic error falls from 2.514 to 0.202 and from 0.626 to 0.149, respectively. An additional audit retains incomplete and invalid submissions in a coverage-sensitive analysis. Observed symbolic-accuracy gains are less consistent across modules, and random acquisition is competitive with disagreement scoring. Measurement savings occur in every module, but unequal batch constraints prevent attributing them solely to acquisition quality. These results support the complete protocol as a promising measurement-efficient configuration, while leaving its causal components and generalization beyond noiseless direct-equation tasks unresolved.
\end{abstract}
\input{main_body}

\input{disclosures}
\clearpage
\begingroup\small
\setlength{\bibsep}{4pt}
\interlinepenalty=10000
\bibliographystyle{unsrtnat}
\bibliography{references}
\endgroup
\clearpage
\input{supplement_numbering}
\section*{Supplementary Material S1}
\input{supplement_body}

\end{document}

%% file: main_body.tex
\section{Introduction}
Scientific problem solving couples two decisions: which observation to obtain next and which explanation to retain afterward. A discovery agent can waste measurements by repeatedly testing similar inputs, or fit a familiar equation before establishing its dependencies. Improving these decisions requires an evaluation that distinguishes the measurements collected, the predictive quality of the resulting function, and whether that function expresses the hidden law. These objectives are related, but success on one does not guarantee success on the others.

For example, a function may closely predict the outputs observed in one input range while omitting a weak dependency that becomes important elsewhere. Conversely, an agent may identify the right functional structure with imperfect fitted constants. Reporting measurement counts, numerical fit, and symbolic recovery together makes these different outcomes visible and helps locate where a discovery procedure succeeds or fails.

NewtonBench makes this problem accessible through hidden physics simulators whose governing laws deliberately depart from textbook equations \cite{zheng2026newtonbench}. Altered exponents and dependencies reduce the usefulness of memorized formulas. Its experiments also associate some failures of interpreter-enabled agents with premature exploitation of early hypotheses. This motivates inspecting how acquisition and equation construction are organized, rather than treating access to a numerical tool as sufficient for discovery. Our study tests a protocol within NewtonBench's existing task suite.

Equation discovery has a substantial history outside large language models (LLMs). Evolutionary search \cite{schmidt2009distilling}, sparse identification of dynamics \cite{brunton2016sindy}, AI Feynman \cite{udrescu2020aifeynman}, and PySR \cite{cranmer2023pysr} recover interpretable mathematical structure; SRBench compares symbolic-regression methods systematically \cite{lacava2021srbench}. Although many evaluations use fixed datasets, discovery is not inherently passive: Bongard and Lipson combined model inference with informative perturbation experiments \cite{bongard2007automated}. LLM-SR incorporates language-model proposals into equation search \cite{shojaee2024llmsr}, while LLM-SRBench examines scientific equation discovery on memorization-resistant tasks \cite{shojaee2025llmsrbench}. DiscoveryWorld and BoxingGym broaden agent evaluation to simulated discovery and experimental design \cite{jansen2024discoveryworld,gandhi2025boxinggym}.

Our acquisition rules draw on the principle that competing hypotheses help identify informative measurements. Bayesian experimental design formalizes information gathering \cite{chaloner1995bayesian}; query by committee selects through disagreement \cite{seung1992query}; Bayesian Active Learning by Disagreement uses a probabilistic information criterion \cite{houlsby2011bald}. BED-LLM connects language-model information gathering with Bayesian experimental design \cite{choudhury2025bedllm}. Here the committees are language-model proposals, not calibrated posterior samples, and the scores are heuristic disagreement measures rather than mutual-information estimates.

We evaluate an \emph{explore-then-commit} configuration. Structured probes and automatic numerical diagnostics support initial inference. A language model proposes equations, a programmatic planner selects further inputs, and a fresh prompt synthesizes a final law from fixed observations and any draft. Interpreter-enabled variants additionally permit model-initiated Python analysis. This allocation of work relates to program-aided language models and Program of Thoughts \cite{gao2023pal,chen2023pot}, and to separating LLM proposals from external checks in robotic planning \cite{pesjak2026robot}. Unlike verification against a known planning model, our numerical checks cannot certify an unknown law's symbolic correctness.

The study asks whether the complete configuration improves measurement use and recovery relative to free-form agents, whether disagreement scoring improves on random acquisition, and whether the findings persist across models and stricter numerical accounting. Across 576 trials, the clearest pattern is fewer measurements and lower mean numerical error. Symbolic gains are concentrated in fewer modules, and random selection remains competitive. We also identify submissions whose finite reported error conceals incomplete predictions. The contribution is therefore a reproducible protocol comparison and evaluation audit: it establishes the behavior of the implemented configurations while distinguishing that evidence from causal claims about their individual components.

\section{Materials and Methods}
\label{sec:methods}
\subsection{Tasks and Experimental Grid}
Each task hides a scalar function $y=f^\star(x)$. An agent queries inputs and outputs, then submits an executable Python function named \texttt{discovered\_law}. The twelve modules cover gravity, Coulomb and magnetic forces, Fourier conduction, Snell refraction, radioactive decay, harmonic motion, Malus intensity, sound speed, Hooke energy, Bose--Einstein occupation, and heat transfer. We use the direct-output \texttt{vanilla\_equation} tier, law version \texttt{v0}, and zero observation noise. The labels denote benchmark task families; the altered laws are not claims about actual physics.

GPT-4.1-mini supplies $8\times12\times2\times2=384$ trials across eight configurations, twelve modules, easy and medium difficulties, and two rollouts per cell. GPT-4.1 supplies another $8\times12\times2=192$ medium trials. This second grid was selected after larger primary-study differences appeared on medium tasks; it is a targeted replication. Recorded aliases are \texttt{gpt41mini} and \texttt{gpt41}; provider routes and exact hosted-model snapshots are not fully resolved. Generation temperature is 0.4 and planner pool seeds are 0 and 1. Rollouts are not paired through common model-randomness seeds. All 576 final submissions are retained, including sixteen interpreter-enabled baseline outcomes that reached their turn limit; no terminal failure artifacts occur.

\subsection{Configurations, Budgets, and Protocol}
We compare two free-form baselines, without and with an interpreter, against variance, pairwise, and random planners in both interpreter conditions. Interpreter access means the LLM can execute its own analysis code. Every planner also receives automatic sensitivity estimates and power-law regression diagnostics, which the baselines do not receive. Thus an agent without an interpreter can still receive substantive numerical assistance.

Baselines permit ten model-action rounds plus a forced final-submission call when needed. Their experiment parser accepts variable-length measurement lists; a suggested batch limit of twenty is unenforced. Planner protocol \texttt{p4} allows eight exploration turns followed by one synthesis turn without an interpreter or at most three with one. Each post-bootstrap planner batch is capped at four measurements. An exploration turn can instead perform analysis, repair an action, or submit a draft. Turn limits therefore do not specify equal measurement budgets. We count individual simulator evaluations, and interpret savings as outcomes of these complete interaction rules.

\begin{figure}[htbp]
\begin{adjustwidth}{-\extralength}{0cm}
\centering\includegraphics[width=\linewidth]{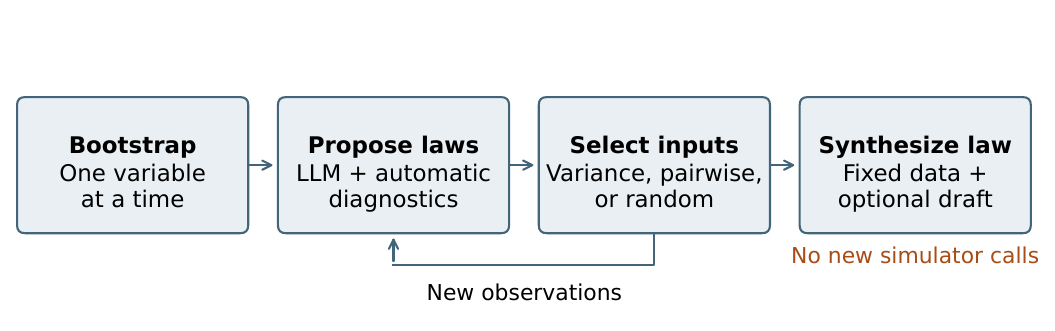}
\caption{The planner protocol. The large language model (LLM) proposes laws; every variant receives automatic numerical diagnostics. Fresh synthesis receives fixed observations and any draft, with no new simulator access.}
\label{fig:protocol}
\end{adjustwidth}
\end{figure}

Figure~\ref{fig:protocol} shows the division of work. Bootstrap probes vary one coordinate at a time around a transformed midpoint, using at least three and at least $d+1$ measurements for $d$ inputs. Positive observations support log-ratio sensitivity estimates and a least-squares fit of $\log y$ to the log inputs. The supplied exponents and residuals can expose multiplicative structure directly. During exploration, the model proposes up to five hypotheses. Scored planners require at least two parseable candidates; the random variant can continue with fallback hypotheses. A draft submission ends exploration.

Scored acquisition evaluates calibrated hypotheses on a pool of 2048 scrambled Sobol inputs. A fit filter retains candidates consistent with observations. Variance acquisition measures disagreement after standardizing each hypothesis across the pool; pairwise acquisition uses mean absolute disagreement after median and interquartile-range normalization. Boundary penalties and within-batch diversity modify selection. Random acquisition samples unused pool points without replacement, bypassing filtering and score regularization while sharing bootstrap probes, diagnostics, batch constraints, and synthesis. Supplementary Sections S1--S3 give pseudocode, formulas, constants, and domains.

Calibration fits a scale and, with sufficient observations, an offset to each candidate. It therefore adjusts a proposed equation's predictions without optimizing every nuisance parameter independently. The diagnostics and fit filter use only acquired observations, and the shared evaluation grid is never presented to the agent. Candidate-pool coordinates follow each simulator's linear or logarithmic input distribution. These choices define the tested acquisition procedure; no pool-size or calibration ablation was conducted.

Fresh synthesis receives the task, observations, and any nontrivial draft rather than the full exploration history. It permits local analysis when an interpreter is available and refuses new experiment requests. A valid synthesis law replaces the draft; otherwise the draft is retained. The instruction favors retaining a draft unless observations disagree, so the reset does not eliminate all anchoring. Recorded traces confirm a synthesis pass and nonempty diagnostics in all 432 planner trials; three use bootstrap measurements alone.

\subsection{Metrics, Aggregation, and Verification}
Symbolic Accuracy (SA) is NewtonBench's saved binary GPT-4.1 judgment at temperature 0.6. The evaluator can accept fitted-constant differences while rejecting incorrect dependencies. SA is benchmark-judged recovery, not deterministic symbolic equivalence. We preserve every recorded judgment without rerunning the evaluator.

Numerical error is replayed locally on deterministic, shared grids of 5000 inputs per module, using difficulty-specific truth. The benchmark's magnitude-based root-mean-squared logarithmic error (RMSLE) is
\begin{equation}
 R=\sqrt{\frac{1}{|V|}\sum_{j\in V}\left[\log(1+|\hat f(x_j)|)-\log(1+|f^\star(x_j)|)\right]^2}.
\label{eq:rmsle}
\end{equation}
The implementation removes input NaN pairs, then averages non-NaN squared log differences indexed by $V$. Infinite terms can invalidate the score, and function exceptions invalidate evaluation. Absolute values discard sign. Because the retained set $V$ can depend on the submission, a finite score can conceal undefined predictions. Reported RMSLE means exclude non-finite trial scores and disclose the retained count.

Our retrospective coverage analysis requires a submission to return finite predictions at every sampled input where the target is finite. Let $C_t$ indicate this completeness for trial $t$. We report $Q(\tau)$, the fraction of \emph{all} trials with $C_t=1$ and finite RMSLE at most $\tau$. Invalid and incomplete submissions remain in the denominator. Thresholds 0.01, 0.1, and 1 summarize the curve; they are descriptive choices, not calibrated scientific tolerances. Sign disagreements are audited separately.

Each baseline row has 24 rollouts; each planner-family row averages three equally represented variants, totaling 72. It does not select the best variant or combine answers. Balanced cells give modules equal weight for SA and measurement means, although finite-only error averages can weight them unequally. We average repeats and variants within modules for sensitivity analysis, using 10,000 module-resampling draws, leave-one-module-out ranges, and four secondary sign-flip diagnostics (Supplementary Section S4). These fixed-suite summaries do not resolve uncertainty from two rollouts per cell.

Numerical replay uses independent evaluation inputs shared across models, configurations, and repeats. The difficulty-specific truth changes while the input grid stays fixed. This removes unequal test samples as a source of between-configuration error differences; it does not make differently acquired training observations equivalent or establish performance outside the benchmark's input distribution.

Supplementary Material S1, appended after the references, documents additional methods and results. The S2 code-and-data snapshot is provided with the arXiv ancillary files and contains sanitized records, code, dependencies, and hashes. Offline replay reproduces the existing scores and leaves the original data unchanged. All 49 protocol, replay, and coverage tests pass. The reference dependency snapshot uses Python 3.10.18; this audit uses Python 3.11.7. Claude and Codex assisted software and manuscript development, including this revision's analysis, writing, and data-rendered figures; details appear in the software-assistance disclosure. No new model-generated trial outcomes were created for this revision.

\section{Results}
\subsection{Measurement Use and Numerical Fit}
\begin{figure}[htbp]
\begin{adjustwidth}{-\extralength}{0cm}
\centering\includegraphics[width=\linewidth]{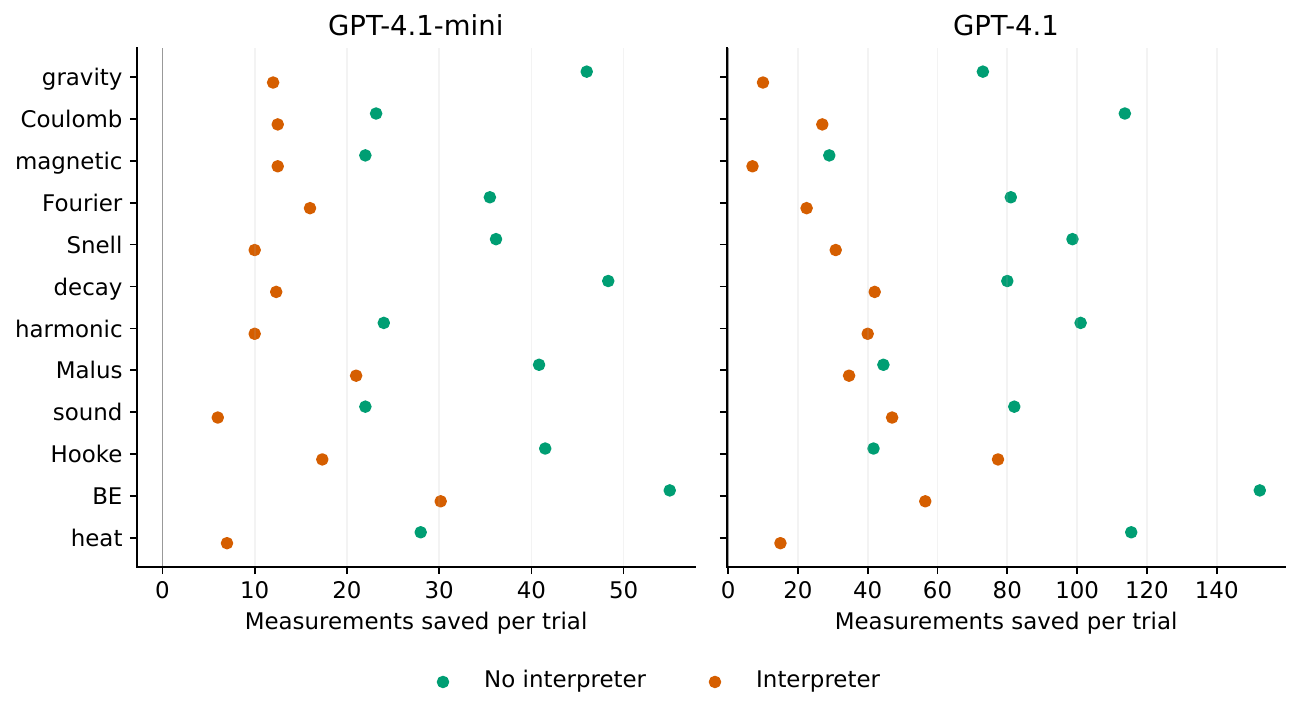}
\caption{Medium-task measurement savings. Each point subtracts six planner rollouts' mean from two matching baseline rollouts' mean. All module-level differences are positive; comparisons include unequal batch limits and automatic planner diagnostics.}
\label{fig:measurements}
\end{adjustwidth}
\end{figure}

Table~\ref{tab:family} summarizes every model--difficulty--family condition. On medium tasks, interpreter-enabled planners use 8.6 measurements per rollout versus 22.5 for the GPT-4.1-mini baseline, a 61.9\% reduction from unrounded means. GPT-4.1 uses 8.9 versus 43.0, a 79.3\% reduction. Without an interpreter, savings are 79.9\% and 88.6\%, respectively. Figure~\ref{fig:measurements} shows positive savings in all twelve modules in each of the four model--tool comparisons. This consistency describes the implemented protocols, including their unequal batch constraints.

\begin{table}[htbp]
\begin{adjustwidth}{-\extralength}{0cm}
\caption{Complete family-level results. N/C denote no interpreter/interpreter access; GPT prefixes are omitted. SA denotes Symbolic Accuracy and RMSLE root-mean-squared logarithmic error. Credited counts symbolic recoveries; Finite counts finite RMSLE scores; Meas. counts simulator evaluations per rollout. RMSLE averages only finite scores. Planner rows average all three acquisition variants.}
\label{tab:family}\centering\small\input{tables/family.tex}
\end{adjustwidth}
\end{table}

Medium-task mean RMSLE for interpreter-enabled planners falls from 2.514 to 0.202 on GPT-4.1-mini and from 0.626 to 0.149 on GPT-4.1. Without an interpreter, the changes are 2.059 to 1.167 and 3.018 to 0.634. Thus both numerical fit and measurement use favor the planner families in all four comparisons. The second model reproduces this direction despite stronger baseline recovery, rather than merely reproducing an advantage over a uniformly unsuccessful baseline.

Easy GPT-4.1-mini tasks show less separation in symbolic recovery: interpreter-enabled planners match the baseline at 54.2\%, while using 8.5 versus 20.8 measurements. The no-interpreter planner family reaches 47.2\% versus 20.8\% for its baseline. Supplementary Table S4 reports all eight easy-task configurations, including variants below their matching baseline. Lower measurement counts do not establish lower total compute or API cost; recorded output tokens omit prompt tokens and the protocol adds external calculations and synthesis.

The family average also conceals different easy-task tradeoffs. With an interpreter, variance receives 11 symbolic credits, compared with 13 for baseline and 14 each for pairwise and random. Random has the lowest mean numerical error in this group, at 0.102. Thus the unchanged family-level SA does not indicate identical behavior among its components, and the medium-task ordering of scored and random numerical performance does not extend to every condition.

\subsection{Symbolic Recovery and the Random Control}
On GPT-4.1-mini medium tasks, planner-family SA increases from 8.3\% to 37.5\% without an interpreter and from 16.7\% to 40.3\% with one. These gains are heterogeneous: the no-interpreter comparison has five positive module effects and seven ties, while the interpreter comparison has four positive effects, one negative effect, and seven ties. GPT-4.1 gains are smaller, at 12.5 and 5.6 percentage points; the latter involves two positive modules and ten ties. Mean differences remain positive when any one module is removed, but none of the four unadjusted sign-flip diagnostics is below 0.05 (Supplementary Section S4.1 and Table S2).

For GPT-4.1 without an interpreter, three module effects are positive, one is negative, and eight are tied. The second model therefore repeats the positive aggregate direction without producing a broad shift in symbolic success across the suite. A tied module can reflect common success or common failure; the recovery heatmap distinguishes these cases instead of treating every tie as evidence that the task was solved.

\begin{table}[htbp]
\begin{adjustwidth}{-\extralength}{0cm}
\caption{All medium-task configurations; 24 rollouts per row. Int. denotes interpreter access. Credited, Finite, and Complete are counts out of 24. Complete requires predictions throughout the sampled finite-target domain; RMSLE means exclude non-finite scores.}
\label{tab:backends}\centering\small\input{tables/backends.tex}
\end{adjustwidth}
\end{table}

The complete acquisition comparison in Table~\ref{tab:backends} gives no consistent symbolic-recovery ordering. For GPT-4.1-mini with an interpreter, variance and pairwise each recover 10 of 24 laws versus 9 for random. Without an interpreter, random obtains 10 recoveries, pairwise 9, and variance 8. GPT-4.1 with an interpreter reverses the first ordering: random obtains 11 recoveries versus 10 for each scored variant. Differences of one recovery cannot establish reliable separation with only two rollouts per module.

Scored interpreter-enabled variants do have lower mean RMSLE than random in both models: 0.189 and 0.155 versus 0.263 on GPT-4.1-mini, and 0.115 and 0.118 versus 0.213 on GPT-4.1. This ordering is descriptive; filtering and fallback behavior differ alongside scoring. Supplementary Figure S1 shows the module-level recovery pattern. Seven medium GPT-4.1-mini modules remain unrecovered by any configuration; harmonic motion, Malus intensity, and Bose--Einstein occupation have no credited recoveries anywhere in the collection. GPT-4.1 does obtain some Snell recoveries.

\subsection{Prediction Completeness and Error}
Fifteen of 576 replayed scores are non-finite. Another fourteen submissions have finite RMSLE despite missing predictions where the target is finite; all fourteen concern Snell refraction. One easy GPT-4.1-mini interpreter-enabled baseline predicts at only 770 of 4259 finite-target points yet receives RMSLE 0.957. The Snell target itself is finite on 4259 easy and 4264 medium points out of 5000, so completeness is assessed relative to that domain, not undefined target outputs. In total, 29 submissions fail the completeness check. Separately, two complete submissions have sign disagreements that magnitude RMSLE cannot penalize as sign errors.

\begin{figure}[htbp]
\begin{adjustwidth}{-\extralength}{0cm}
\centering\includegraphics[width=\linewidth]{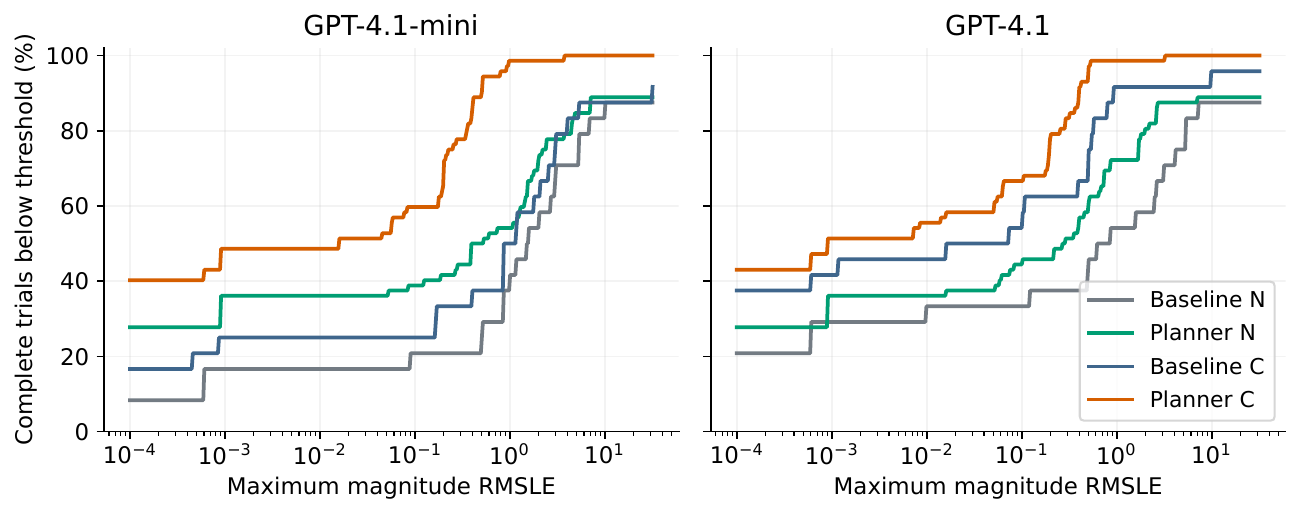}
\caption{Coverage-sensitive medium-task error curves. Each curve counts all rollouts, requiring complete predictions on the sampled finite-target domain and RMSLE at most the threshold. Incomplete or invalid submissions prevent curves from reaching 100\%.}
\label{fig:coverage}
\end{adjustwidth}
\end{figure}

The fit advantage persists under this stricter accounting (Figure~\ref{fig:coverage}; Supplementary Table S3). All 72 interpreter-enabled medium planner submissions are complete in each model. At $\tau=0.1$, the all-trial rates are 59.7\% versus 25.0\% for the GPT-4.1-mini baseline and 66.7\% versus 54.2\% for GPT-4.1. The gaps depend on the tolerance. Requiring completeness prevents selective prediction domains from improving the reported pass rate, while retaining the distinction between a close predictive function and a recovered symbolic law.

Both sign-disagreeing submissions have RMSLE below 0.1 and consequently satisfy that coverage-sensitive threshold. Completeness repairs one weakness of finite-only averages, but does not repair the underlying metric's sign insensitivity. The analysis therefore preserves the legacy error measure for comparability while exposing the additional information needed to interpret it. No original judge decision or submitted function is replaced by the retrospective audit.

\section{Discussion}
The repeated finding is a useful complete configuration: structured acquisition, automatic diagnostics, and fresh synthesis produce lower mean numerical error using fewer measurements under the implemented rules. The numerical advantage survives an audit that retains incomplete and invalid submissions. This matters for scientific agents whose measurements are costly, but the practical interpretation should remain specific to the interface: lower simulator use is established, whereas superior information acquisition at an equal measurement budget is not.

The largest limitation is that several interventions change together. Baselines accept uncapped measurement arrays; planners use bounded batches, bootstrap probes, automatic regression, hypothesis prompting, and fresh synthesis. In particular, log-linear diagnostics can supply much of a multiplicative law's structure, even without model-initiated Python. A cap-matched baseline is the first decisive follow-up. Separately removing diagnostics and fresh synthesis, with matched measurement availability and model-call opportunities, would identify their contributions. Fitting a dedicated symbolic-regression solver to the same observations would test the value added by LLM synthesis. These comparisons have not been performed.

Random acquisition further limits a scoring-specific explanation. Its competitive recovery suggests that shared protocol components may account for much of the observed improvement; it does not establish equivalence of acquisition rules. More independent rollouts and prospectively fixed comparisons are needed. Module-resampling summaries cannot compensate for two repetitions per cell, and leave-one-module-out positivity cannot exclude effects concentrated in a few tasks. Medium-only replication was selected after inspecting primary results and uses the same provider's model family. Neither design feature establishes broad external validity.

Evaluation also separates predictive usefulness from scientific explanation. A harmonic submission can have RMSLE below $10^{-3}$ despite using the wrong damping dependency (Supplementary Section S4.4). Close numerical fit can miss weak dependencies; sampled completeness cannot guarantee off-grid validity. The stochastic GPT-4.1 judge shares a model family with discovery, can accept coefficient differences, and has not been independently adjudicated. The sign-insensitive metric and limited provider metadata further constrain interpretation and exact recollection reproducibility.

The study covers noiseless direct equations, not harder laws, alternative versions, or indirect dynamics. Sensitivity to pool size, bootstrap design, regularization, and temperature remains unmeasured. Both tested baselines benefit from interpreter access, so these data do not demonstrate reversal of NewtonBench's tool-assistance paradox. No preregistration is documented; the coverage and leave-one-module-out analyses are retrospective. Within that scope, the evidence supports making acquisition and synthesis separately inspectable, accounting for all numerical assistance, and reporting predictive coverage alongside symbolic recovery.

For subsequent evaluations, this suggests publishing the observations available at synthesis and the numerical diagnostics supplied to each configuration. Those artifacts make the division of work auditable and allow later comparisons to reuse identical evidence rather than recollecting it under different conditions.

\section{Conclusions}
Across 576 NewtonBench trials, the explore-then-commit configuration uses fewer measurements and achieves lower mean numerical error than the implemented baselines. The fit advantage repeats across two models and remains favorable when incomplete predictions are retained in the analysis. Symbolic gains are less uniform, and random acquisition competes with disagreement scoring. These findings support the complete protocol within the tested conditions. A measurement-cap-matched baseline, followed by diagnostic and synthesis ablations with more independent rollouts, is needed to determine which components cause the improvements and how reliably they generalize.

%% file: tables/family.tex
\begin{tabular}{llrrrrrr}
\toprule
Model / difficulty & Configuration & $n$ & Credited & SA (\%) & RMSLE & Finite & Meas. \\
\midrule
4.1-mini / easy & Baseline N & 24 & 5/24 & 20.8 & 1.405 & 24/24 & 43.2 \\
4.1-mini / easy & Planner N & 72 & 34/72 & 47.2 & 1.165 & 71/72 & 9.4 \\
4.1-mini / easy & Baseline C & 24 & 13/24 & 54.2 & 1.046 & 24/24 & 20.8 \\
4.1-mini / easy & Planner C & 72 & 39/72 & 54.2 & 0.290 & 71/72 & 8.5 \\
4.1-mini / medium & Baseline N & 24 & 2/24 & 8.3 & 2.059 & 23/24 & 44.0 \\
4.1-mini / medium & Planner N & 72 & 27/72 & 37.5 & 1.167 & 66/72 & 8.8 \\
4.1-mini / medium & Baseline C & 24 & 4/24 & 16.7 & 2.514 & 24/24 & 22.5 \\
4.1-mini / medium & Planner C & 72 & 29/72 & 40.3 & 0.202 & 72/72 & 8.6 \\
4.1 / medium & Baseline N & 24 & 4/24 & 16.7 & 3.018 & 22/24 & 95.2 \\
4.1 / medium & Planner N & 72 & 21/72 & 29.2 & 0.634 & 68/72 & 10.9 \\
4.1 / medium & Baseline C & 24 & 9/24 & 37.5 & 0.626 & 24/24 & 43.0 \\
4.1 / medium & Planner C & 72 & 31/72 & 43.1 & 0.149 & 72/72 & 8.9 \\
\bottomrule
\end{tabular}

%% file: tables/backends.tex
\begin{tabular}{llrrrrrr}
\toprule
Model & Configuration & Credited & SA (\%) & RMSLE & Finite & Complete & Meas. \\
\midrule
4.1-mini & Baseline, no int. & 2 & 8.3 & 2.059 & 23 & 21 & 44.0 \\
4.1-mini & Baseline, int. & 4 & 16.7 & 2.514 & 24 & 22 & 22.5 \\
4.1-mini & Variance, no int. & 8 & 33.3 & 1.394 & 22 & 22 & 9.0 \\
4.1-mini & Pairwise, no int. & 9 & 37.5 & 0.899 & 21 & 21 & 7.7 \\
4.1-mini & Random, no int. & 10 & 41.7 & 1.195 & 23 & 21 & 9.8 \\
4.1-mini & Variance, int. & 10 & 41.7 & 0.189 & 24 & 24 & 8.3 \\
4.1-mini & Pairwise, int. & 10 & 41.7 & 0.155 & 24 & 24 & 8.3 \\
4.1-mini & Random, int. & 9 & 37.5 & 0.263 & 24 & 24 & 9.0 \\
4.1 & Baseline, no int. & 4 & 16.7 & 3.018 & 22 & 22 & 95.2 \\
4.1 & Baseline, int. & 9 & 37.5 & 0.626 & 24 & 23 & 43.0 \\
4.1 & Variance, no int. & 7 & 29.2 & 0.504 & 22 & 22 & 10.7 \\
4.1 & Pairwise, no int. & 7 & 29.2 & 0.532 & 24 & 22 & 11.3 \\
4.1 & Random, no int. & 7 & 29.2 & 0.875 & 22 & 20 & 10.7 \\
4.1 & Variance, int. & 10 & 41.7 & 0.115 & 24 & 24 & 9.2 \\
4.1 & Pairwise, int. & 10 & 41.7 & 0.118 & 24 & 24 & 8.3 \\
4.1 & Random, int. & 11 & 45.8 & 0.213 & 24 & 24 & 9.2 \\
\bottomrule
\end{tabular}

%% file: disclosures.tex
\section*{Reproducibility and Software Assistance}
Supplementary Material S1 follows the references. The S2 code-and-data snapshot
is included in the arXiv ancillary files and provides the 576 recorded trials,
offline replay, analysis scripts, dependencies, and a SHA-256 manifest. Original
NewtonBench software is distributed under its retained MIT license. Model
provider calls and symbolic judging are not required for offline reproduction.

Anthropic Claude and OpenAI Codex assisted with software implementation,
manuscript development, literature checking, analysis code, and figure-generation
code. Earlier tool-version identifiers were not fully preserved. The figures
are rendered from the recorded data. No additional model-generated experimental
outcomes were created for this revision.

%% file: supplement_numbering.tex
\setcounter{section}{0}
\setcounter{table}{0}
\setcounter{figure}{0}
\setcounter{equation}{0}
\setcounter{algorithm}{0}
\renewcommand{\thesection}{S\arabic{section}}
\renewcommand{\thesubsection}{\thesection.\arabic{subsection}}
\renewcommand{\thetable}{S\arabic{table}}
\renewcommand{\thefigure}{S\arabic{figure}}
\renewcommand{\theequation}{S\arabic{equation}}
\renewcommand{\thealgorithm}{S\arabic{algorithm}}
\renewcommand{\theHsection}{supp.\arabic{section}}
\renewcommand{\theHtable}{supp.\arabic{table}}
\renewcommand{\theHfigure}{supp.\arabic{figure}}
\renewcommand{\theHequation}{supp.\arabic{equation}}
\providecommand{\theHalgorithm}{\arabic{algorithm}}
\renewcommand{\theHalgorithm}{supp.\arabic{algorithm}}

%% file: supplement_body.tex
This supplementary material accompanies the NewtonBench protocol study and supplies detailed methods and secondary results. LLM denotes large language model, SA Symbolic Accuracy, and RMSLE root-mean-squared logarithmic error. All values derive from the same 576 recorded trials; no additional acquisition or judging was performed. The companion code-and-data archive is designated S2. 

\section{Exploration and Synthesis Protocol}
\label{sec:protocol}
A configured bootstrap count of three is raised to at least $d+1$ for a $d$-variable task. Structured probes begin at the midpoint in transformed coordinates, then move each variable separately to its 0.75 fraction. If more probes are needed, the 0.25 fraction is used; thus one-variable Hooke tasks receive three probes. All other inputs remain centered. The system computes log-ratio sensitivity estimates where positive observations permit them, and fits
\begin{equation}
 \log y \approx \beta_0 + \sum_{k=1}^{d}\beta_k\log x_k
\end{equation}
by least squares when sufficient positive data are available. The estimates, residuals, and fitted exponents are presented as advisory diagnostics, not as ground truth. They can directly expose a power-law structure and therefore constitute substantive numerical assistance, including for the agents without an interpreter.

During exploration, a model may propose up to five symbolic hypotheses in a JSON committee. Scored planners require at least two parseable hypotheses; otherwise they request repair. The random variant can continue using fallback hypotheses because selection does not require a disagreement signal. Interpreter-enabled variants may instead perform a Python calculation. A draft submission ends exploration and is retained for the synthesis stage.

After exploration, a new message history receives the task description, serialized observations, and any nontrivial draft. The preceding conversation is not copied, although the draft carries information derived during exploration. The synthesis instruction favors retaining a draft unless the observations disagree; a context reset therefore does not eliminate all anchoring. Interpreter-enabled synthesis permits local Python analysis. Experiment requests are refused. A valid synthesis submission replaces the draft; the draft is retained only when synthesis yields no valid submission. The audit of recorded traces verifies one synthesis pass in each of the 432 planner trials. Only three planner trials use bootstrap measurements alone. All 432 record nonempty automatic diagnostics.

\begin{algorithm}[htbp]
\caption{Planner protocol evaluated in this study}
\label{alg:protocol}
\begin{algorithmic}[1]
\State Construct candidate pool $\mathcal X$; collect structured bootstrap data $\mathcal D$.
\For{at most eight exploration turns}
 \State Present observations and automatic diagnostics to the LLM.
 \State Receive a draft law, an allowed local calculation, or a hypothesis committee.
 \If{a draft law is submitted}
   \State Store draft; \textbf{break}.
 \ElsIf{an allowed local calculation is requested}
   \State Execute and return its output; \textbf{continue}.
 \ElsIf{a scored planner lacks a valid committee}
   \State Request repair; \textbf{continue}.
 \EndIf
 \State Select up to four unused inputs: uniformly for random, otherwise by filtered disagreement and diversity.
 \State Query simulator and append the measurements to $\mathcal D$.
\EndFor
\State Start fresh synthesis with $\mathcal D$ and any draft; disable new measurements.
\State Return a valid synthesis law, or the draft if synthesis yields none.
\end{algorithmic}
\end{algorithm}
\section{Acquisition Details}
\subsection{Candidate Pool, Filtering, and Scores}
The candidate pool contains $2048=2^{11}$ scrambled Sobol points mapped into module-specific linear or logarithmic input coordinates. The power-of-two size is compatible with the Sobol construction and bounds scoring cost; it was not selected by a reported pool-size sweep. The coordinates and bounds are specified in Section~\ref{app:domains}. Previously selected pool points are excluded from further acquisition. Structured bootstrap probes are constructed separately rather than sampled uniformly from the pool.

For scored acquisition, parseable hypotheses are calibrated to observed data using $\tilde h_i(x)=a_i h_i(x)+b_i$. With at least three finite pairs, both coefficients are fitted by least squares; with two pairs only a scale is fitted. The filter ranks candidates by root-mean-square residual divided by median absolute observed output, using a mean-absolute-output or unit fallback if that denominator is not positive and finite. Up to four candidates with normalized error at most 0.75 survive. If fewer than two meet the threshold, the two lowest-error candidates survive; if fewer than two fits are valid, filtering falls back to the original committee. Common nuisance symbols are initially evaluated as unit constants before this affine calibration. Consequently, the filter does not optimize every free parameter in an arbitrary equation.

Let $p_{ij}=\tilde h_i(x_j)$ denote a retained prediction on the currently available pool. Variance acquisition standardizes each hypothesis across pool points using its finite-prediction mean $\mu_i$ and population standard deviation $\sigma_i$:
\begin{equation}
 z_{ij}=(p_{ij}-\mu_i)/\sigma_i,\qquad
 s_{\mathrm{var}}(x_j)=\frac{1}{|I_j|}\sum_{i\in I_j}(z_{ij}-\bar z_j)^2,
\end{equation}
where $I_j$ indexes finite standardized predictions and $\bar z_j$ is their mean. Constant finite hypotheses are represented by zeros. At least two usable predictions are needed. This is variance across pool-standardized hypotheses, not variance divided by the mean output at an input point.

Pairwise acquisition centers each hypothesis by its finite-prediction median $m_i$ and scales it by its interquartile range $q_i$, using its standard deviation if $q_i=0$. A nonpositive fallback scale makes that hypothesis unusable. With normalized predictions $r_{ij}$,
\begin{equation}
 s_{\mathrm{pair}}(x_j)=\frac{1}{|P_j|}\sum_{(i,k)\in P_j}|r_{ij}-r_{kj}|,
\end{equation}
where $P_j$ contains finite pairs with $i<k$. Thus pairwise acquisition uses normalized mean absolute disagreement rather than raw absolute output differences.

Both scores are rank-transformed, penalized near input-domain boundaries, weighted by the fraction of finite predictions, and combined with diversity during batch selection. Section~\ref{app:selection} gives the constants and selection formula. Random acquisition samples unused pool points without replacement and bypasses fit filtering and score regularization. It shares the bootstrap, diagnostics, exploration prompt, batch constraint, and fresh synthesis. It therefore compares acquisition configurations within a common protocol but is not a clean ablation of phase separation.
\subsection{Regularization and Batch Diversity}
\label{app:selection}
Let $\rho_j\in[0,1]$ be the ordinal rank of a finite raw score, with ranks assigned in ascending score order. Ties are assigned positions by the implementation's sorting routine rather than averaged ranks. Let $u_{jk}$ be candidate $j$'s percentile position along coordinate $k$, and $e_j=\min_k\min(u_{jk},1-u_{jk})$. The regularized score is
\begin{equation}
 \tilde s_j=\rho_j\left[\operatorname{clip}\left(\frac{e_j-0.08}{0.5-0.08},0,1\right)\right]^{1.5}v_j,
\end{equation}
where $v_j$ is the fraction of finite hypothesis predictions. Rank power is one. Candidates within the highest-scoring quarter are eligible, with at least the batch size eligible when enough finite scores exist. The first selected point has highest normalized eligible score. Subsequent points maximize $0.65s'_j+0.35\delta_j$, where $s'_j$ is min--max-normalized eligible score and $\delta_j$ is distance to the nearest point already selected in that batch in percentile coordinates, divided by $\sqrt d$. This is a heuristic batch-diversity rule, not a joint information-gain optimizer. Random acquisition bypasses these operations.
\section{Input Domains and Evaluation Grids}
\label{app:domains}
Table~\ref{tab:domains} lists the input distributions used for both candidate-pool mapping and numerical evaluation. The acquisition pool uses scrambled Sobol inputs; the evaluation grid uses an independently seeded NumPy PCG64 generator. Grid seeds are obtained from the first eight bytes of the SHA-256 digest of a version identifier and module name, interpreted as a little-endian unsigned integer. The 5000-point grid is reused across models, methods, and difficulties, while ground truth remains difficulty-specific. Full seeds, bounds, source hashes, and grid hashes are included in Archive S2. These generated test points are not sent to the agent.

\begin{table}[htbp]
\begin{adjustwidth}{-\extralength}{0cm}
\caption{Input domains. L denotes log-uniform sampling and U uniform sampling. Units and parameter interpretation follow the source simulator; these are computational benchmark domains.}
\label{tab:domains}\centering\small
\begin{tabular}{lp{0.73\linewidth}}\toprule
Module & Variables and distributions\\\midrule
Gravity & $m_1,m_2:\mathrm L[1,10^3]$; $r:\mathrm L[1,10]$\\
Coulomb & $q_1,q_2,r:\mathrm L[10^{-1},10]$\\
Magnetic & $I_1,I_2,r:\mathrm L[10^{-3},10^{-1}]$\\
Fourier & $k:\mathrm L[10^{-1},10]$; $A:\mathrm L[10^{-4},10^{-2}]$; $\Delta T:\mathrm L[10,10^3]$; $d:\mathrm L[10^{-2},1]$\\
Snell & $n_1,n_2:\mathrm U[1,1.5]$; angle in degrees: $\mathrm U[0,90]$\\
Decay & $N_0:\mathrm L[1,10^2]$; $\lambda:\mathrm L[10^{-3},10^{-1}]$; $t:\mathrm L[10^{-2},10]$\\
Harmonic & $k:\mathrm L[10^2,10^4]$; $m:\mathrm L[10^{-1},10]$; $b:\mathrm L[10^{-2},1]$\\
Malus & $I_0:\mathrm L[10^2,2\times10^3]$; angle in radians: $\mathrm U[10^{-6},\pi/2]$\\
Sound & $\gamma:\mathrm U[1.3,1.7]$; $T:\mathrm L[10,10^3]$; $M:\mathrm L[10^{-3},10^{-1}]$\\
Hooke & $x:\mathrm L[10^{-3},1]$\\
Bose--Einstein & $\omega:\mathrm L[10^8,10^{10}]$; $T:\mathrm L[10,10^3]$\\
Heat transfer & $m:\mathrm L[10^{-3},10^3]$; $c:\mathrm L[10^2,10^4]$; $\Delta T:\mathrm L[10,10^3]$\\\bottomrule
\end{tabular}
\end{adjustwidth}
\end{table}
\section{Extended Results and Sensitivity Analyses}
\subsection{Module-Level Recovery and Uncertainty}
Each baseline row contains 24 rollouts. Each planner-family row averages all three acquisition variants with equal representation, giving 72 rollouts; it is neither a best-of-three selection nor an ensemble that combines their answers. SA and measurement means give every module equal weight because their cells are balanced. Finite-only RMSLE means can implicitly give modules different weights when submissions are non-finite, which motivates reporting coverage and all-trial curves as well.

For module-level comparisons, average repeats and planner variants within each module, then subtract the matching baseline. We report the twelve individual effect directions, their mean, a 95\% percentile interval obtained by resampling the twelve module effects 10,000 times with seed zero, and the range after removing each module in turn. These are sensitivity summaries for a fixed benchmark suite. They do not estimate uncertainty over an independently sampled task population and do not resolve rollout-level uncertainty. A two-sided sign-flip enumeration is retained for the four medium-task SA contrasts as a secondary diagnostic, with its unadjusted values disclosed. We do not use trial-index paired tests or homogeneous-binomial confidence intervals as evidence of generalization. Sign-flip enumeration assumes independent, sign-symmetric module effects under the null. The study does not randomize whole-protocol labels at module level, so these values are sensitivity diagnostics rather than design-based randomization evidence.

On GPT-4.1-mini medium tasks, the planner families raise observed SA from 8.3\% to 37.5\% without an interpreter and from 16.7\% to 40.3\% with one. The average differences remain positive when any one module is removed (Table~\ref{tab:module}). However, seven modules have zero family--baseline SA difference in each comparison. The interpreter-enabled comparison has four positive module effects and one negative effect, while the no-interpreter comparison has five positive effects. A positive mean does not describe uniform symbolic improvement.

\begin{table}[htbp]
\begin{adjustwidth}{-\extralength}{0cm}
\caption{Medium-task Symbolic Accuracy (SA) differences in percentage points (pp). $+/-/0$ counts modules with positive, negative, and tied planner--baseline effects. Resampling intervals reweight the twelve fixed modules; leave-one-out ranges remove one module at a time. Neither summarizes stochastic rollout uncertainty or population generalization.}
\label{tab:module}
\centering\small\input{tables/module_effects.tex}
\end{adjustwidth}
\end{table}

On GPT-4.1, the observed gains are 12.5 points without an interpreter and 5.6 with one. The latter is concentrated in two modules, with ten ties. The four unadjusted sign-flip diagnostic values are 0.0625 and 0.125 for GPT-4.1-mini without and with an interpreter, and 0.5 for both GPT-4.1 contrasts. None is below 0.05 even before multiplicity adjustment. Some module-resampling intervals exclude zero, but those intervals are not inversions of the sign-flip calculation. We report the distinction rather than selecting the more favorable summary. More rollouts are needed to determine the stability of the underlying cell means.

\begin{figure}[htbp]
\begin{adjustwidth}{-\extralength}{0cm}
\centering\includegraphics[width=\linewidth]{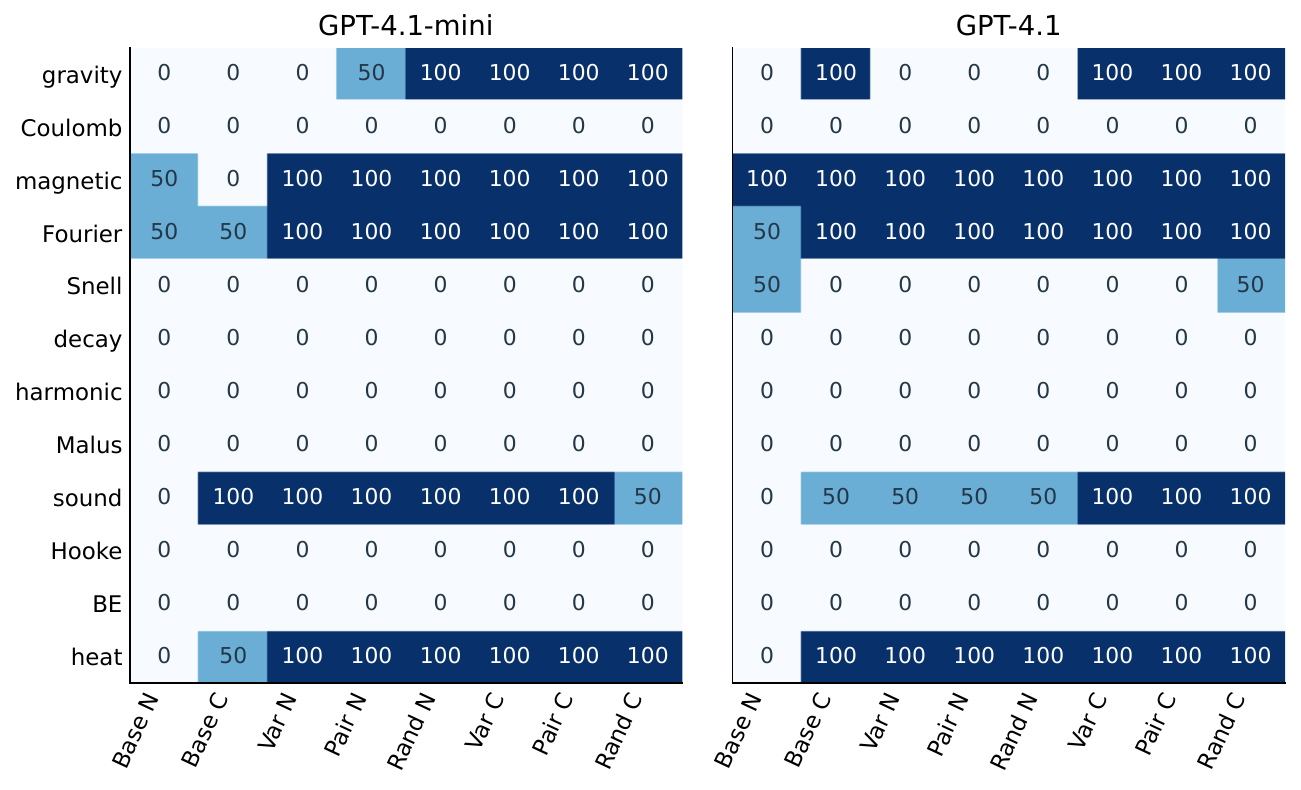}
\caption{Medium-task Symbolic Accuracy percentages for each module and configuration. Each cell contains two rollouts, so a single changed outcome moves it by 50 points. N and C denote no interpreter and interpreter access; Var, Pair, and Rand denote acquisition rules. Colors encode the displayed percentages, not confidence.}
\label{fig:recovery}
\end{adjustwidth}
\end{figure}

Figure~\ref{fig:recovery} shows where symbolic credit occurs. Medium-task GPT-4.1-mini planners recover several altered algebraic laws, particularly gravity with an interpreter, magnetic force, Fourier conduction, and heat transfer. Seven modules remain unrecovered by any configuration at that difficulty. Across the entire collection, underdamped harmonic motion, Malus's law, and Bose--Einstein occupation have no credited recoveries. GPT-4.1 produces some Snell recoveries, so the latter is not a universal failure across the two models.

\subsection{Prediction-Domain Audit}
\label{sec:coverage-results}
The numerical metric is the benchmark's magnitude-based RMSLE, replayed on shared grids. For trial $t$, let $G_t$ contain the sampled inputs with finite ground-truth outputs and let $C_t=1$ only when the submitted function evaluates successfully and predicts finite outputs everywhere in this nonempty set. The coverage-sensitive pass rate is
\begin{equation}
 Q(\tau)=\frac{1}{n}\sum_{t=1}^{n}\mathbf{1}\{C_t=1,\ R_t\text{ finite},\ R_t\le\tau\}.
\end{equation}
All trials remain in the denominator. Completeness is assessed on the sampled finite-target domain, not between grid points, and does not check output sign. The thresholds are retrospective descriptive choices.

The replay reproduces all previously reported shared-grid scores. Fifteen of 576 trial scores are non-finite. In addition, fourteen submissions have finite RMSLE but omit predictions at some points where the target is finite. All fourteen are Snell-law submissions. For example, one GPT-4.1-mini easy baseline with an interpreter predicts on only 770 of 4259 finite-target points, yet receives a finite RMSLE of 0.957. A low or moderate reported score can therefore refer to a restricted subset of the evaluation domain.

This is partly distinct from invalid target points: the Snell truth is finite on 4259 easy and 4264 medium inputs out of 5000, reflecting its restricted domain. Our completeness check is relative to those finite-target points, not a demand to predict a real-valued output where the target itself is undefined. Fourteen finite-score submissions fail that additional check. Together with the fifteen non-finite-score outcomes, 29 submissions fail the complete-domain criterion. Two submissions also disagree with the target's sign at one or more finite pairs, an error the magnitude metric does not penalize as a sign error.

\begin{table}[htbp]
\begin{adjustwidth}{-\extralength}{0cm}
\caption{Coverage-sensitive medium-task results. Complete gives submissions defined on the whole sampled finite-target domain. The remaining columns give $100Q(\tau)$: the percentage of \emph{all} rollouts that are complete and have finite RMSLE at most $\tau$. Non-finite and incomplete predictions remain in the denominator. Thresholds are retrospective descriptive choices.}
\label{tab:coverage}
\centering\small\input{tables/coverage.tex}
\end{adjustwidth}
\end{table}

The numerical-fit pattern remains favorable under this stricter accounting (Table~\ref{tab:coverage}). All 72 medium-task interpreter-enabled planner submissions have complete finite-target coverage in each model. At $\tau=0.1$, their all-trial rates are 59.7\% versus 25.0\% for the GPT-4.1-mini baseline and 66.7\% versus 54.2\% for GPT-4.1. The gaps vary with the threshold, and are descriptive rather than claims about a uniquely appropriate tolerance. Complete coverage still does not imply a correct equation.

\subsection{Complete Easy-Task Comparison}
Table~\ref{tab:easy} reports every GPT-4.1-mini easy configuration using the same shared-grid replay and completeness criterion as the main paper. Every row retains all 24 rollouts for SA and measurement means. RMSLE means include only finite scores, whose counts are shown. No GPT-4.1 easy grid was collected. This table adds a presentation of existing records rather than new experiments or a selection of favorable variants.

\begin{table}[htbp]
\begin{adjustwidth}{-\extralength}{0cm}
\caption{All GPT-4.1-mini easy-task configurations, 24 rollouts per row. Int. denotes interpreter access. Credited, Finite, and Complete are counts out of 24. Complete requires predictions throughout the sampled finite-target domain. RMSLE means exclude non-finite scores.}
\label{tab:easy}\centering\small\input{tables/easy_backends.tex}
\end{adjustwidth}
\end{table}

\subsection{Weak Dependencies and Numerical Fit}
\label{sec:weak}
An instructive harmonic submission uses $\sqrt{k/m-(b/(2m))^2}$ where the target is $\sqrt{k/m-b/(2m^2)}$. The damping dependencies differ algebraically, but their numerical effect is weak across much of the sampled domain. A recorded submission consequently has RMSLE below $10^{-3}$ while receiving no symbolic credit. This is evidence of metric insensitivity to a weak dependency, not evidence that the judge rejected an equivalent rearrangement. Automatic power-law diagnostics can help expose multiplicative structure, but a good fit to that structure cannot certify additive, trigonometric, or reciprocal-exponential dependencies.

Sign disagreement is a separate failure mode. Both audited sign-disagreeing submissions have complete sampled domains: the GPT-4.1 medium Bose--Einstein variance planner with an interpreter, trial 0, and the GPT-4.1-mini easy decay random planner with an interpreter, trial 1. Their magnitude RMSLE values are approximately 0.0512 and 0.0970, respectively. Both therefore pass the coverage-sensitive threshold of 0.1 despite wrong signs at some inputs. These outcomes are distinct from the 29 incomplete submissions and remain in all reported analyses.
\section{Artifact Verification}
The S2 code-and-data snapshot includes the original 43-test offline suite, six additional coverage tests, and scripts to regenerate the new analysis. Source-data hashes are checked before and after law execution. The coverage audit instruments the numerical metric to count finite-target and finite-prediction pairs without changing the value returned by the metric. When evaluation fails before that calculation, the submission is recorded as incomplete. The raw judge outcome is never replaced by a coverage result. The supplementary validation additionally checks all 576 row identities, every regenerated RMSLE value against the earlier replay, figure/table inputs, and preservation of the source snapshot. Consult the archive README for executable commands and the distinction between original results, retrospective analyses, and prospective experiments that have not been run.

\subsection{Portable Files and Offline Commands}
Archive S2 includes sanitized copies of recorded trial JSON, the flat trial table, original benchmark and planner code, deterministic replay, and the retrospective analyses. Copied paths and organization identifiers are replaced; submitted laws, recorded evaluations, measurement counts, and output-token counts are verified unchanged. A SHA-256 manifest covers the shipped files. Replaying records never invokes the acquisition agents or the symbolic judge. It executes the saved law functions locally against the source simulator's ground truth.

From the extracted archive root, the executable commands are:
\begin{quote}\small\ttfamily
python3 -m unittest tests/test\_planner\_study.py \textbackslash\\
\hspace*{1em}paper/test\_shared\_grid\_rmsle.py \textbackslash\\
\hspace*{1em}paper\_make\_20260908/analysis/test\_audit.py
\end{quote}
The three test paths belong to one command. Then run
\begin{quote}\small\ttfamily
python3 paper\_make\_20260908/analysis/audit.py\\
python3 paper\_make\_20260908/analysis/assets.py
\end{quote}
The first script regenerates coverage, shared-grid, trace, and source-integrity records; the second regenerates figures and tables, including the complete easy-task comparison. All outputs stay inside the new manuscript directory. The six added tests cover partial NaN predictions, invalid target points, infinite predictions, sign changes, empty finite-target sets, and mismatched array shapes.

The arXiv source archive supplies the main manuscript and this supplementary material in a single document, together with their bibliography and graphics. It uses the standard LaTeX article class. The top-level file is \texttt{main.tex}; compile it with XeLaTeX and BibTeX, or with Tectonic. The concise main paper preserves the earlier numerical results; the coverage and leave-one-module-out analyses are retrospective. Measurement-cap matching, component ablations, and further rollouts are prospective studies and are not included as completed results.

%% file: tables/module_effects.tex
\begin{tabular}{llrrrr}
\toprule
Model & Tool access & $\Delta$ SA (pp) & Resampling interval & $+/-/0$ & Leave-one-out range \\
\midrule
4.1-mini & No interpreter & +29.2 & [+8.3, +54.2] & 5/0/7 & [+22.7, +31.8] \\
4.1-mini & Interpreter & +23.6 & [+2.8, +47.2] & 4/1/7 & [+16.7, +27.3] \\
4.1 & No interpreter & +12.5 & [-8.3, +33.3] & 3/1/8 & [+4.5, +18.2] \\
4.1 & Interpreter & +5.6 & [+0.0, +15.3] & 2/0/10 & [+1.5, +6.1] \\
\bottomrule
\end{tabular}

%% file: tables/coverage.tex
\begin{tabular}{llrrrr}
\toprule
Model & Configuration & Complete & $\tau=0.01$ & $\tau=0.1$ & $\tau=1$ \\
\midrule
4.1-mini & Baseline N & 21/24 & 16.7 & 20.8 & 41.7 \\
4.1-mini & Planner N & 64/72 & 36.1 & 38.9 & 54.2 \\
4.1-mini & Baseline C & 22/24 & 25.0 & 25.0 & 50.0 \\
4.1-mini & Planner C & 72/72 & 48.6 & 59.7 & 98.6 \\
4.1 & Baseline N & 22/24 & 33.3 & 33.3 & 54.2 \\
4.1 & Planner N & 64/72 & 36.1 & 44.4 & 72.2 \\
4.1 & Baseline C & 23/24 & 45.8 & 54.2 & 91.7 \\
4.1 & Planner C & 72/72 & 55.6 & 66.7 & 98.6 \\
\bottomrule
\end{tabular}

%% file: tables/easy_backends.tex
\begin{tabular}{lrrrrrr}
\toprule
Configuration & Credited & SA (\%) & RMSLE & Finite & Complete & Meas. \\
\midrule
Baseline, no int. & 5 & 20.8 & 1.405 & 24 & 24 & 43.2 \\
Baseline, int. & 13 & 54.2 & 1.046 & 24 & 22 & 20.8 \\
Variance, no int. & 10 & 41.7 & 1.105 & 24 & 24 & 9.5 \\
Pairwise, no int. & 12 & 50.0 & 1.109 & 23 & 23 & 9.3 \\
Random, no int. & 12 & 50.0 & 1.279 & 24 & 24 & 9.5 \\
Variance, int. & 11 & 45.8 & 0.392 & 23 & 23 & 8.0 \\
Pairwise, int. & 14 & 58.3 & 0.381 & 24 & 23 & 8.2 \\
Random, int. & 14 & 58.3 & 0.102 & 24 & 24 & 9.3 \\
\bottomrule
\end{tabular}